\documentclass{coowa}

\usepackage[toc,page,header]{appendix}
\usepackage{comment}

\usepackage{algorithm}
\usepackage{algpseudocode}

\usepackage{xspace}
\usepackage{amsmath}
\usepackage{amssymb}
\usepackage{amsfonts}
\usepackage{mathtools}

\usepackage{adjustbox}
\usepackage{accsupp}
\usepackage{transparent}

\usepackage{bbding}
\usepackage{wrapfig}

\crefname{algorithm}{Algorithm}{Algorithms}
\Crefname{algorithm}{Algorithm}{Algorithms}

\newcommand{\done}[1]{\textcolor{red}{[DONE]}}

\definecolor{ShortText}{RGB}{144,216,212}
\definecolor{LongText}{RGB}{128,200,195}
\definecolor{Visual}{RGB}{210,180,220}
\definecolor{bellow}{RGB}{229,181,89}
\definecolor{lightblue}{RGB}{230,240,250}

\newcommand{\bagel}{%
  \IfFileExists{sections/figures/bagel.png}{%
    \raisebox{-0.2ex}{\includegraphics[height=1em]{sections/figures/bagel.png}}%
  }{%
    \textcolor{bellow}{O}%
  }%
}

\title{The DAWN of World-Action Interactive Models}

\author[1,2,*]{Hongbo Lu}
\author[1,*]{Liang Yao}
\author[1,*]{Chenghao He}
\author[1,*]{Haoyu Wang}
\author[3]{Xiang Gu}
\author[1,2]{Xianfei Li}
\author[1,\dagger]{Wenlong Liao}
\author[1]{Tao He}
\author[1,\dagger,\ddagger]{Pai Peng}

\affiliation[1]{COWARobot Co. Ltd}
\affiliation[2]{Shanghai Jiao Tong University}
\affiliation[3]{Hohai University}

\contribution[*]{Equal Contribution}
\contribution[\dagger]{Corresponding Author}
\contribution[\ddagger]{Project Lead}

\abstract{
A plausible scene evolution depends on the maneuver being considered, while a good maneuver depends on how the scene may evolve. Existing World Action Models (WAMs) largely miss this reciprocity, treating world prediction and action generation as either isolated parallel branches or rigid predict-then-plan pipelines. We formalize this perspective as World-Action Interactive Models (WAIMs), and instantiate it in autonomous driving with \textbf{DAWN} (\textbf{D}enoising \textbf{A}ctions and \textbf{W}orld i\textbf{N}teractive model), a simple yet strong latent generative baseline. DAWN operates in a compact semantic latent space and couples a \emph{World Predictor} with a \emph{World-Conditioned Action Denoiser}: the predicted world hypothesis conditions action denoising, while the denoised action hypothesis is fed back to update the world prediction, so that both are recursively refined during inference. Rather than eliminating test-time world evolution altogether or rolling out the full future in pixel space, DAWN performs a short explicit latent rollout that is sufficient to support long-horizon trajectory generation in complex interactive scenes. Experiments show that DAWN achieves strong planning performance and favorable safety-related results across multiple autonomous driving benchmarks. More broadly, our results suggest that interactive world-action generation is a principled path toward truly actionable world models.
}

\date{\today}

\checkdata[Project Page]{\url{https://cowarobot-ai.github.io/}}

\correspondence{\email{volans.liao@cowarobot.com}, \email{pengpai@cowarobot.com}}

\begin{document}

\maketitle

%

\section{Introduction}
\label{sec:intro}

World models~\cite{feng2025survey,ding2025understanding} aim to predict how the environment may evolve. World Action Models (WAMs)~\cite{ye2026world,cen2025worldvla,yuan2026fast} extend this idea to decision-making by modeling future world evolution together with the agent's actions. 
To be actionable, a WAM should predict different futures for different actions.
This requirement is especially pronounced in autonomous driving~\cite{tang2026causalvad,wang2025comdrive,yang2025uncad,zhou2026opendrivevla,han2025dme}, where the future relevant to decision making is inherently action-contingent: whether a gap remains feasible, whether another agent yields, and which interactions become safety-critical all depend on the ego maneuver under consideration. For planning, the objective is not to predict a passive future of the scene, but to infer a future that is physically plausible under candidate actions and informative for choosing among them. Therefore, we argue that a useful WAM should not merely represent world and action together, but it should let them co-evolve during inference.

\begin{figure*}[t]
    \centering
    \includegraphics[width=0.99\linewidth]{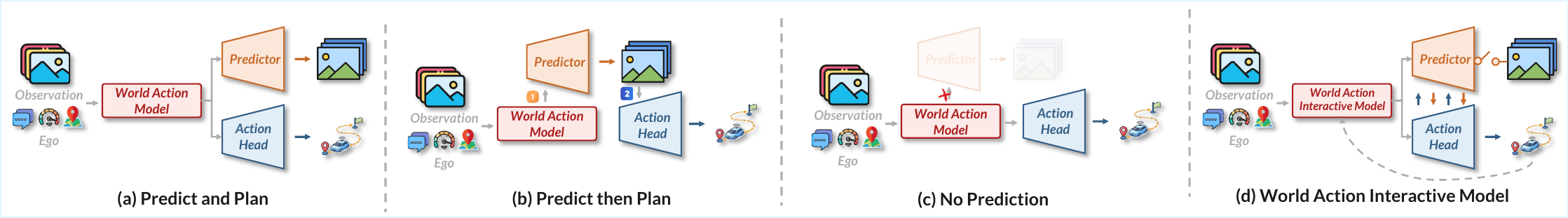}
    \caption{From WAMs to WAIM. Existing WAMs typically predict world and action in parallel, sequentially, or without explicit test-time rollout. In contrast, WAIM keeps a short latent world rollout and recursively couples world prediction with action generation during inference.}
    \label{teaser}
\end{figure*}

As illustrated in Fig.~\ref{teaser}, existing World Action Models are still largely built around a structural decoupling between world generation and action generation. A common design is to predict future world states and actions in parallel from shared visual context, using separate heads or branches for scene evolution and motion planning~\cite{zhang2025epona,ye2026world,bartoccioni2025vavim}. Another common design is a sequential pipeline: first forecast future observations, occupancy, or latent scene states, and then plan actions on top of these predicted futures~\cite{li2025imagidrive,zhao2025forecasting}. Although these strategies may improve representation sharing or planning accuracy, they still treat one side as fixed with respect to the other at generation time. Parallel designs allow world and action to be correlated, but not to iteratively reshape one another; sequential designs condition action on a frozen future hypothesis, rather than a future that evolves together with the action hypothesis. As a result, they fall short of modeling the bidirectional, action-dependent nature of decision-relevant futures in interactive driving.

Recent works such as Fast-WAM~\cite{yuan2026fast} suggest that explicit world rollout is not always necessary at inference time. In relatively simple domains, world modeling can mainly serve as a training signal, while test-time action generation reduces to a direct policy interface. We view such zero-rollout inference as one endpoint of a broader design space rather than a universal solution. In complex interactive scenes, some explicit future evolution remains useful for reasoning about moving agents and obstacles. Importantly, this rollout need not span the full task horizon or operate in pixel space: a model can generate long-horizon actions while rolling out the world only over a shorter latent horizon. This places inference-time rollout in WAMs on a continuum, ranging from zero-rollout methods such as Fast-WAM to full predict-then-plan models.

To move beyond structural decoupling and the binary choice between full rollout and no rollout, we advocate World-Action Interactive Models (WAIMs). WAIMs treat future world states and actions as coupled variables inferred together during generation, rather than as independent outputs or stages in a fixed one-way pipeline. As illustrated in Fig.~\ref{teaser}(d), the current world hypothesis refines the action hypothesis, while the emerging action hypothesis feeds back to revise the predicted world evolution, forming a coherent future-action pair. This is the sense in which WAIMs are interactive: not merely bidirectional information flow inside the architecture, but an inference process where world and action hypotheses co-evolve. This distinction matters whenever the decision-relevant future depends on the action being formed, rather than on scene dynamics alone. Therefore, a WAIM does not first predict a world and then act in it.  Instead, it jointly infers a future in which world evolution and decision making remain mutually aligned. 

In this work, we instantiate WAIM for autonomous driving with \textbf{DAWN} (\textbf{D}enoising \textbf{A}ctions and \textbf{W}orld i\textbf{N}teractive model), a latent generative model that operates in a compact semantic space and avoids expensive pixel-level future rendering. Rather than eliminating inference-time world evolution or rolling out the world over the full planning horizon, DAWN uses a short explicit latent rollout to support long-horizon action generation in complex interactive scenes. Concretely, DAWN couples a World Predictor with a World-Conditioned Action Denoiser: the predicted world hypothesis conditions action denoising, while the denoised action hypothesis is fed back to update the world rollout. Through this recursive interaction, DAWN allows world and action hypotheses to co-evolve during generation, providing a minimal instantiation of the WAIM principle.

Experiments on several autonomous driving benchmarks validate that DAWN achieves strong overall planning performance and favorable safety-oriented results. On NAVSIM v1, for example, DAWN achieves the best perception-free PDMS of 89.1 and obtains the best Time-to-Collision score, which is consistent with our goal of making action generation more aware of future world evolution. These results suggest that interactive world-action generation provides a practical path toward safer and more actionable driving models.

Our contributions are summarized as follows:
\begin{itemize}
    \item We identify action-contingent reciprocity as the missing principle in existing WAMs and formulate World-Action Interactive Models.
    \item We introduce DAWN, a short-rollout latent architecture that couples world prediction and action denoising through recursive interaction.
    \item We achieve remarkable perception-free planning on representative benchmarks, demonstrating significant improvements in trajectory accuracy and interactive safety.
\end{itemize}

\section{Methodology}
\subsection{Problem Formulation}
\label{sec:problem_formulation}

We consider policy learning from a current observation $o$ and a task instruction $l$. Let $a_{1:H}$ denote an action chunk over horizon $H$, and let $v_{1:T}$ denote a future world representation over horizon $T$, e.g., future observations or latent future states. A standard policy directly models
\begin{equation}
p(a_{1:H} \mid o, l).
\end{equation}

A \emph{World Action Model} (WAM) extends this formulation by explicitly introducing the future world as an intermediate variable and modeling
\begin{equation}
p(v_{1:T}, a_{1:H} \mid o, l).
\end{equation}
Equivalently, the action distribution is obtained by marginalizing over possible futures:
\begin{equation}
p(a_{1:H} \mid o, l)
=
\int p(v_{1:T}, a_{1:H} \mid o, l)\, dv_{1:T}.
\end{equation}

We define a \emph{World-Action Interactive Model} (WAIM) as a special class of WAMs in which future world and future action are inferred as coupled variables rather than generated independently or in a fixed one-way order. Formally, WAIM seeks a self-consistent pair $(\hat{v}_{1:T}, \hat{a}_{1:H})$ such that
\begin{equation}
\hat{v}_{1:T} = F_{\theta}(o, l, \hat{a}_{1:H}),
\qquad
\hat{a}_{1:H} = G_{\phi}(o, l, \hat{v}_{1:T}),
\end{equation}
which in practice can be realized through iterative interaction:
\begin{equation}
(v_{1:T}^{(k+1)}, a_{1:H}^{(k+1)})
=
\mathcal{I}_{\Theta}(v_{1:T}^{(k)}, a_{1:H}^{(k)}; o, l).
\end{equation}
Thus, the key distinction is that a WAM jointly models future world and action, while a WAIM jointly infers them through interaction.

\begin{figure*}[t]
    \centering
    \includegraphics[width=0.99\linewidth]{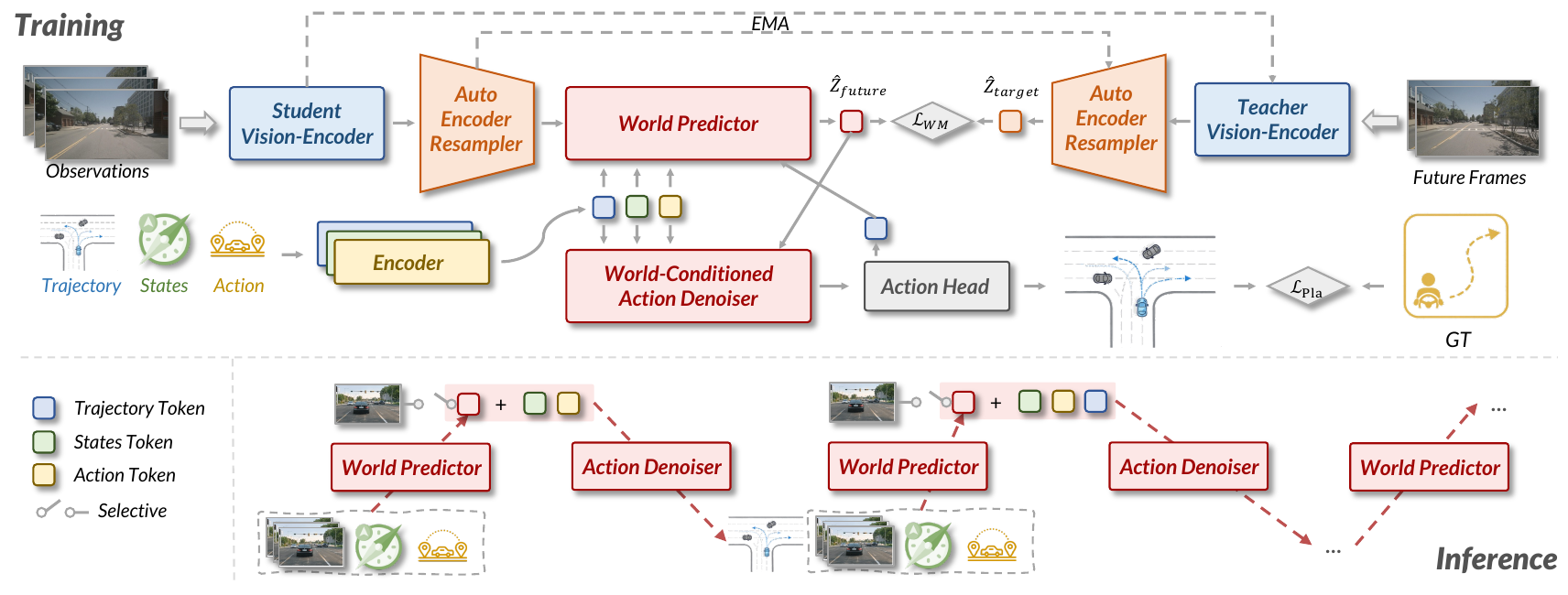}
    \caption{Overview of DAWN. During training, DAWN learns compact latent world tokens with a Student/Teacher Vision-Encoder pair and an Auto-Encoder Resampler, supervises short latent rollout with a World Predictor, and trains a World-Conditioned Action Denoiser for trajectory generation. During inference, the Action Denoiser initializes actions from resampler latents and then recursively refines them with predictor rollouts. This couples world prediction and action generation in latent space without pixel-space future rendering.}
    \label{overview}
\end{figure*}

\subsection{DAWN Architecture}
\label{sec:architecture}

DAWN instantiates WAIM with an interactive world-action architecture in latent space.
As shown in Fig.~\ref{overview}, it consists of a Student Vision-Encoder, a training-time Teacher Vision-Encoder, an Auto Encoder Resampler, a World Predictor, a World-Conditioned Action Denoiser, and a lightweight Action Head.

Given the current observation $o$, the Student Vision-Encoder extracts dense visual tokens
\begin{equation}
u = E_{\mathrm{stu}}(o).
\end{equation}
In our implementation, both the student and teacher branches use V-JEPA 2 Large~\cite{assran2025v} as the vision backbone. Since the dense encoder tokens are expensive to roll out directly, we compress them with an Auto Encoder Resampler, which is a learned bottleneck autoencoder operating in token space:
\begin{equation}
z = R_{\mathrm{stu}}(u).
\end{equation}
This yields a compact latent world representation for downstream interaction. During training, future observations $o^{+}$ are processed by the Teacher Vision-Encoder and its corresponding resampler to produce target future latents
\begin{equation}
z_{\mathrm{target}} = R_{\mathrm{tea}}(E_{\mathrm{tea}}(o^{+})),
\end{equation}
which supervise the world modeling branch. The teacher branch is only used during training.

The core of DAWN is the recursive interaction between a World Predictor and a World-Conditioned Action Denoiser. The World Predictor is implemented as a causal Transformer that predicts future latent world tokens from the current latent context and the current action hypothesis. The World-Conditioned Action Denoiser is implemented as a DiT, which denoises action tokens conditioned on both the latent context and the predicted future world. 
Let $c$ denote the encoded condition tokens, including ego-state and high-level action or route tokens. The Action Denoiser additionally receives role-specific queries that indicate whether it is producing an initial proposal or refining an action using a predictor rollout. DAWN performs
\begin{equation}
a_{1:H}^{(0)} = G_{\phi}(q_{\mathrm{prop}}, c, z),
\qquad
z_{\mathrm{future}}^{(r)} = P_{\theta}(z, c, a_{1:H}^{(r)}),
\qquad
a_{1:H}^{(r+1)} = G_{\phi}(q_{\mathrm{ref}}^{(r)}, c, z_{\mathrm{future}}^{(r)}, a_{1:H}^{(r)}).
\end{equation}
Here $q_{\mathrm{prop}}$ and $q_{\mathrm{ref}}^{(r)}$ are role-specific query embeddings for proposal generation and refinement. The denoiser weights are shared across both roles; only the input source and query embeddings differ.

After the final interaction step, the denoised action states are decoded by the Action Head into the final trajectory prediction. Notably, DAWN does not require rolling out the full action horizon in world space: the world branch only needs to evolve a short latent future that is sufficient to support long-horizon action generation. In this way, DAWN forms a self-consistent world-action hypothesis through iterative interaction, while avoiding expensive pixel-space future rendering.

\subsection{Training}
\label{sec:training}


\noindent\textbf{Stage 1. Vision pretraining.}
We first pretrain the Student Vision-Encoder on large-scale driving video data, including \textsc{OpenScene}~\cite{peng2023openscene}, \textsc{DrivingDojo}~\cite{wang2024drivingdojo}, and \textsc{CoVLA}~\cite{arai2025covla}. All datasets are converted into a unified video format and sampled with a sliding \texttt{window\_stride}. Pretraining is performed at a resolution of $256\times512$ and a frame rate of $2$\,Hz, providing a strong visual prior for downstream latent world modeling.

\noindent\textbf{Stage 2. Auto-Encoder Resampler training.}
Starting from the pretrained encoder, we train the Auto-Encoder Resampler on the same pretraining corpora. This stage learns a compact token-space bottleneck that compresses dense encoder features into latent world tokens while preserving the information required for future prediction and action generation.

\noindent\textbf{Stage 3. World Predictor training.}
We then attach the World Predictor and train it on downstream task datasets such as \textsc{nuScenes}~\cite{caesar2020nuscenes} and \textsc{navsim}~\cite{dauner2024navsim}. In this stage, the predictor learns to roll out task-relevant future latent world states from the compact latent context produced by the pretrained encoder and resampler.

\noindent\textbf{Stage 4. Joint world-action training.}
Finally, we initialize the World Predictor from Stage 3, attach the World-Conditioned Action Denoiser and the Action Head, and jointly train the world and action branches on the target datasets. At this stage, both the predictor and the action denoiser are optimized together. The Action Denoiser is trained in two roles with shared weights: it first generates an initial proposal from the resampler latent context, and then refines the action conditioned on the predictor rollout. Different query and source embeddings specify whether the denoiser is operating in the proposal or interactive refinement role. This training scheme aligns future world rollout and action generation through recursive interaction.

This stage-wise recipe stabilizes optimization and naturally matches the role of each module: large-scale video pretraining provides a strong perceptual prior, the resampler builds an efficient latent bottleneck, the predictor learns future latent evolution, and the final stage turns the model into a full WAIM through coupled world-action training.

\subsection{Inference}
\label{sec:inference}

At inference time, the teacher branch is removed. DAWN first encodes the current observation into a compact latent context
\begin{equation}
z = R_{\mathrm{stu}}(E_{\mathrm{stu}}(o)),
\end{equation}
together with condition tokens $c$ from the non-visual inputs.

Inference follows the same recursive world-action process as training, except that the first action hypothesis can be generated directly from the resampler latent without passing through the World Predictor. Specifically, the World-Conditioned Action Denoiser first produces
\begin{equation}
a_{1:H}^{(0)} = G_{\phi}(q_{\mathrm{init}}, c, z),
\end{equation}
where $q_{\mathrm{init}}$ denotes the initial action queries. DAWN then alternates between short latent world rollout and action denoising:
\begin{equation}
z_{\mathrm{future}}^{(k+1)} = P_{\theta}(z, c, a_{1:H}^{(k)}),
\qquad
a_{1:H}^{(k+1)}
=
G_{\phi}(q_{\mathrm{ref}}^{(k)}, c, z_{\mathrm{future}}^{(k+1)}, a_{1:H}^{(k)}).
\end{equation}
After $K$ refinement steps, the Action Head decodes the final action state into the predicted trajectory,
\begin{equation}
\hat{\tau} = H_{\mathrm{act}}(a_{1:H}^{(K)}).
\end{equation}
A key property of DAWN is that inference supports both \emph{planning from scratch} and \emph{trajectory interactive refinement} within the same architecture. In the first mode, no trajectory prompt is provided, and the model directly predicts $\hat{\tau}$ from $(o,l)$. In the second mode, an initial predicted trajectory can be fed back as an additional prompt for another forward pass, producing a refined trajectory estimate. 

\begin{table}[t]
\caption{Quantitative comparisons on NAVSIM v1 benchmark. The main comparison is conducted against perception-based methods, which share the same planning setting as DAWN. Perception-free methods are included for reference only. DAWN* denotes trained at a resolution of 256$\times$256.}
\label{tab:navsim_v1}
\resizebox{\textwidth}{!}{
\begin{tabular}{l|l|c|cccccc}
\toprule
\textbf{Type}&\textbf{Method} & \textbf{Inputs} & \textbf{NC}$\uparrow$ & \textbf{DAC}$\uparrow$ & \textbf{EP}$\uparrow$ & \textbf{C}$\uparrow$ & \textbf{TTC}$\uparrow$ & \textbf{PDMS}$\uparrow$ \\
\hline
\multirow{8}{*}{Perception-based}&Transfuser~\cite{chitta2022transfuser} & C \& L & 97.7 & 92.8 & 79.2 & 100 & 92.8 & 84.0 \\
&Hydra-MDP~\cite{li2024hydra} & C \& L & 98.4 & 97.7 & 85.0 & 100 & 94.5 & 89.9 \\
&Hydra-MDP++~\cite{li2025hydra} & C \& L & 97.6 & 96.0 & 80.4 & 100 & 93.1 & 86.6 \\
&DiffusionDrive~\cite{liao2025diffusiondrive} & C \& L & 98.2 & 96.2 & 82.2 & 100 & 94.7 & 88.1 \\
&GoalFlow~\cite{xing2025goalflow} & C \& L & 98.4 & 98.3 & 85.0 & 100 & 94.6 & 90.3 \\
&DriveDPO~\cite{shang2025drivedpo} & C \& L & 98.5 & 98.1 & 84.3 & 100 & 94.8 & 90.0 \\
&iPad~\cite{guo2025ipad} & Camera & 99.2 & 97.4 & 87.8 & 99.7 & 96.3 & 91.7 \\
&DriveSuprim~\cite{yao2026drivesuprim} & Camera & 98.6 & 98.6 & 91.3 & 100 & 95.5 & 93.5 \\
\hline
\multirow{6}{*}{Perception-free}&LAW~\cite{li2024enhancing} & C \& L & 97.4 & 93.3 & 78.8 & 100 & 91.9 & 83.8 \\
&World4Drive~\cite{zheng2025world4drive} & C \& L & 97.4 & 94.3 & 79.9 & 100 & 92.8 & 85.1 \\
&Epona~\cite{zhang2025epona} & Camera & 97.9 & 95.1 & 80.4 & 99.9 & 93.8 & 86.2 \\
&Drive-JEPA~\cite{wang2026drive} & Camera & \textbf{98.7} & \textbf{96.2} & 82.9 & 100 & 95.5 & \underline{89.0} \\
\rowcolor{lightblue} 
 & \textbf{DAWN* (Ours)} & Camera & \underline{98.2}& 95.8& \underline{84.2}& 100 & \underline{95.8}&87.9\\
\rowcolor{lightblue} 
 & \textbf{DAWN (Ours)} & Camera &\textbf{98.7}&\underline{95.9}&\textbf{84.3}&100&\textbf{96.0}&\textbf{89.1}\\				
\bottomrule
\end{tabular}
}
\end{table}

\section{Experiments}

In this section, we report the main results of DAWN and conduct ablation studies and further analyses to better understand the advantages of WAIM and the behavior of our model. More detailed results and additional visualizations are provided in the appendix.

\subsection{Experimental Setup}
\label{sec:settings}

\subsubsection{Datasets and Metrics}
\label{sec:datasets_metrics}

We evaluate DAWN on several autonomous driving benchmarks: NAVSIM~\cite{dauner2024navsim} and nuScenes~\cite{caesar2020nuscenes}. NAVSIM evaluates planning quality with simulator-based rule metrics covering collision avoidance, drivable-area compliance, progress, comfort, and time-to-collision, and reports PDMS as the aggregate score. 
On nuScenes, we follow the standard end-to-end planning protocol and report trajectory L2 error and collision rate at 1\,s, 2\,s, and 3\,s, together with their averages. 
For NAVSIM, higher values indicate better performance. For nuScenes, lower L2 error and collision rate are better. Full metric definitions are provided in Appendix~\ref{app:datasets_metrics}.


\subsubsection{Implementation Details}
\label{sec:implementation}

All input videos are sampled at 2 Hz. For the main experiments, inputs are resized/cropped to $512\times256$, while ablation studies are conducted at a lower resolution of $256\times256$ for efficiency. We use V-JEPA 2 Large~\cite{assran2025v} as the vision backbone and compress dense visual tokens with an Auto-Encoder Resampler into compact latent world tokens. The World Predictor is implemented as a causal Transformer, while the World-Conditioned Action Denoiser adopts a DiT-style diffusion backbone and uses 5 sampling steps at inference. Models are trained with bfloat16 mixed precision for 150 epochs, using a peak learning rate of $1\times10^{-4}$, an initial learning rate of $5\times10^{-5}$, 8 warmup epochs, and weight decay 0.04. Large-scale training is conducted on 80 NVIDIA A100 GPUs.


\subsection{Main Results}

We evaluate DAWN on two representative benchmarks and compare it with a range of methods under their respective settings. Additional results and analyses are provided in the appendix.

\noindent\textbf{Results on NAVSIM v1.}
Table~\ref{tab:navsim_v1} reports the NAVSIM v1 results. We mainly compare DAWN with perception-based methods, while listing perception-free results only for reference. Among perception-free models, DAWN achieves the best overall PDMS of 89.1, surpassing Drive-JEPA, while also obtaining the best NC, Ego Progress, and Time-to-Collision scores. This indicates that DAWN is safe, smooth, and sufficiently progressive. Compared with its lower-resolution variant DAWN*, the full model improves PDMS from 87.9 to 89.1, showing the benefit of higher-resolution inputs. Overall, DAWN produces strong planning behavior without relying on an explicit perception stack.

\begin{table}[t]
\centering
\caption{Quantitative comparisons on nuScenes benchmark.}
\begin{tabular}{l|cccc|cccc}
\toprule
\multirow{2}{*}{\textbf{Method}} &
\multicolumn{4}{c|}{\textbf{L2 (m)}$\downarrow$} &
\multicolumn{4}{c}{\textbf{Collision Rate (\%)}$\downarrow$} \\
\cline{2-9}
& 1s & 2s & 3s & \textbf{Avg.} & 1s & 2s & 3s & \textbf{Avg.} \\
\hline
ST-P3~\cite{hu2022st} & 1.33 & 2.11 & 2.90 & 2.11 & 0.23 & 0.62 & 1.27 & 0.71 \\
OccNet~\cite{tong2023scene} & 1.29 & 2.13 & 2.99 & 2.13 & 0.21 & 0.59 & 1.37 & 0.72 \\
UniAD~\cite{hu2023planning} & 0.48 & 0.96 & 1.65 & 1.03 & 0.05 & 0.17 & 0.71 & 0.31 \\
VAD~\cite{jiang2023vad} & 0.41 & 0.70 & 1.05 & 0.72 & 0.07 & 0.18 & 0.43 & 0.23 \\
PPAD~\cite{chen2024ppad} & 0.31 & 0.56 & 0.87 & 0.58 & 0.08 & 0.12 & 0.38 & 0.19 \\
GenAD~\cite{zheng2024genad} & 0.28 & 0.49 & 0.78 & 0.52 & 0.08 & 0.14 & 0.34 & 0.19 \\
\hline
BEV-Planner~\cite{li2024ego} & 0.30 & 0.52 & 0.83 & 0.55 & 0.10 & 0.37 & 1.30 & 0.59 \\
LAW~\cite{li2024enhancing} & 0.26 & 0.57 & 1.01 & 0.61 & 0.14 & 0.21 & 0.54 & 0.30 \\
World4Drive~\cite{zheng2025world4drive} & 0.23 & 0.47 & 0.81 & 0.50 & \underline{0.0}  & 0.12 & \underline{0.33} & 0.16  \\
WorldRFT~\cite{yang2026worldrft} & \underline{0.21} & \underline{0.44} & \underline{0.76} & \underline{0.47} & 0.10 & 0.11 & \textbf{0.23} & \underline{0.15} \\
\hline \rowcolor{lightblue}
\textbf{Ours}  & 	\textbf{0.17}& 	\textbf{0.31}& 	\textbf{0.52}& 	\textbf{0.33}& 	\textbf{0.00}& 	\textbf{0.10}& 	\textbf{0.23}& \textbf{0.11}\\
\bottomrule
\end{tabular}
\label{tab:nuscenes}
\end{table}

\noindent\textbf{Results on nuScenes.} 
Table~\ref{tab:nuscenes} reports end-to-end planning results on the nuScenes benchmark. DAWN achieves state-of-the-art performance across both trajectory accuracy and collision-related metrics. For trajectory prediction, DAWN obtains the lowest L2 error at all horizons, reducing the average L2 error to 0.33\,m, compared with 0.47\,m from the strongest prior method WorldRFT. The gains are especially clear at mid- and long-horizon prediction, where DAWN reduces the 2\,s and 3\,s L2 errors to 0.31\,m and 0.52\,m, respectively. DAWN also achieves the best average collision rate, with leading or tied-leading results across all evaluated horizons. These results show that DAWN improves planning accuracy without sacrificing safety, suggesting that recursive world-action interaction helps produce trajectories that are both precise and collision-aware.



\subsection{Ablation Studies}

\subsubsection{Ablation on Key Components}

\begin{wraptable}{r}{0.36\textwidth}
\centering
\vspace{-0.4cm}
\caption{Component ablation of DAWN. Res., Pre., and Inter. denote the Resampler, Predictor, and interactive update, respectively.}
\label{tab:ablation_components}
\small
\begin{tabular}{ccc|c}
\toprule
\textbf{Res.} &
\textbf{Pred.} &
\textbf{Inter.} &
\textbf{PDMS}$\uparrow$ \\
\midrule
           &             &             & 82.9\\
\checkmark &             &             & 82.8\\
\checkmark & \checkmark  &             & 85.2 \\
\checkmark & \checkmark  & \checkmark  & \textbf{87.9} \\
\bottomrule
\end{tabular}
\vspace{-0.4cm}
\end{wraptable}


We progressively add the key components of DAWN to verify their contributions. This ablation is designed to separate three factors that are otherwise coupled in the full model: compact latent representation, explicit future rollout, and interactive world-action inference. The Auto-Encoder Resampler provides a compact latent world representation, but compression alone does not introduce temporal reasoning. The World Predictor further enables the model to roll out future latent states, providing an explicit future hypothesis for planning. Finally, the interactive design couples the predicted world with action generation, allowing the action hypothesis to be refined according to the evolving world state. As shown in Table~\ref{tab:ablation_components}, adding the Resampler alone does not substantially improve PDMS, indicating that a compact latent bottleneck by itself is not sufficient for stronger planning. Introducing the World Predictor already yields a clear gain, increasing PDMS to 85.2, which suggests that explicit latent future rollout provides useful planning context. Enabling interactive world-action updates further improves PDMS from 85.2 to 87.9. This confirms that the gain does not only come from using a latent world representation or predicting a future world, but also from allowing the world and action hypotheses to refine each other during generation. 

\subsubsection{Ablation on Number of Interactive Rounds}

\begin{wrapfigure}[12]{r}{0.40\columnwidth}
\vspace{-0.35cm}
\centering
\includegraphics[width=0.40\columnwidth]{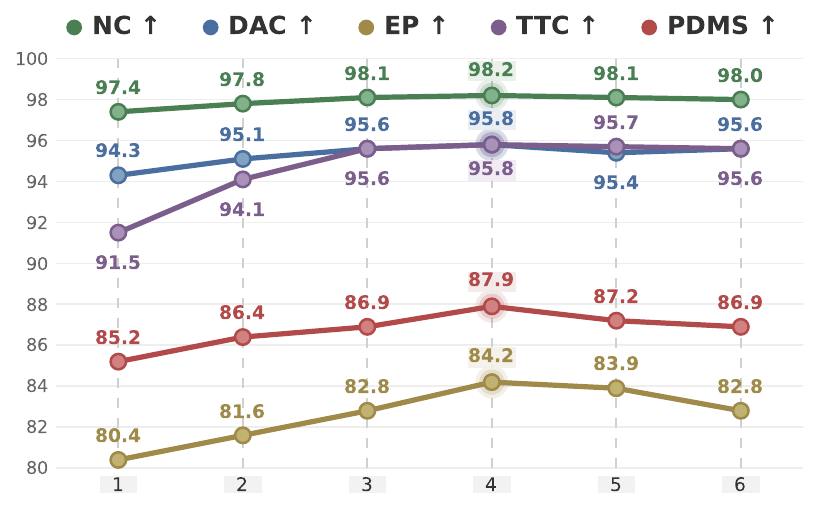}
\caption{Effect of interactive rounds.}
\label{fig:ablation_refine}
\end{wrapfigure}

We further study how iterative refinement affects planning performance. This ablation directly tests whether DAWN benefits from repeated world-action interaction, or whether a single proposal is already sufficient. As shown in Fig.~\ref{fig:ablation_refine}, performance improves steadily as the number of interactive rounds increases from 1 to 4. This trend indicates that each additional round can use the updated latent world hypothesis to further correct the action hypothesis, leading to better progress, time-to-collision, and overall PDMS. After 4 rounds, performance saturates and slightly decreases with additional interactive steps, suggesting that most useful interaction has already been absorbed and further updates provide limited benefit. We therefore use 4 interactive rounds as the default setting in DAWN, which gives the best empirical trade-off between planning quality and inference cost.

\subsubsection{Ablation on Number of Resampler Tokens}

\begin{wraptable}[8]{r}{0.36\textwidth}
\centering
\vspace{-0.35cm}
\caption{Ablation on the number of Resampler output tokens.}
\label{tab:resampler_tokens}
\small
\setlength{\tabcolsep}{4.5pt}
\begin{tabular}{c|c|r}
\toprule
\textbf{\# Tokens} & \textbf{PDMS}$\uparrow$ & \textbf{Lat. (ms)}$\downarrow$\\
\midrule
16 & 82.8 & 331.253\\
64 & 83.2 & 963.645\\
\bottomrule
\end{tabular}
\vspace{-0.3cm}
\end{wraptable}

We also study how the capacity of the Auto-Encoder Resampler affects downstream planning. The resampler controls how much visual information is preserved in the compact latent world representation. As shown in Table~\ref{tab:resampler_tokens}, increasing the number of output tokens from 16 to 64 improves PDMS from 82.8 to 83.2. This suggests that overly aggressive compression may discard planning-relevant scene structure, such as drivable-area geometry, nearby agents, or short-term interaction cues. At the same time, using more latent tokens increases the cost of subsequent world rollout and action denoising. This ablation reflects a capacity-efficiency trade-off in DAWN: the latent bottleneck should be compact enough for efficient rollout, but expressive enough to preserve action-relevant world information.
\subsection{Further Analysis}

\subsubsection{Does World-Action Coupling Really Matter?}

We ablate the two interaction directions in DAWN to test whether its gains come from genuine world-action coupling. Removing World$\rightarrow$Action disables predicted world hypotheses for action denoising, while removing Action$\rightarrow$World makes world rollout independent of the current action hypothesis. Table~\ref{tab:world_action_coupling} shows that full DAWN performs best across all metrics. Removing World$\rightarrow$Action reduces PDMS from 87.9 to 81.6, and removing Action$\rightarrow$World lowers it to 84.9, indicating that either removing world-conditioned action denoising or action-conditioned world rollout weakens the model. These results support the core WAIM principle: world evolution and action generation should mutually constrain each other during inference.

\begin{table}[t]
\centering
\caption{Further analysis on world-action coupling. We remove each direction of interaction to examine whether bidirectional world-action updates are necessary.}
\label{tab:world_action_coupling}
\begin{tabular}{l|cccccc}
\toprule
\textbf{Method} & \textbf{NC}$\uparrow$ & \textbf{DAC}$\uparrow$ & \textbf{EP}$\uparrow$ & \textbf{C}$\uparrow$ & \textbf{TTC}$\uparrow$ & \textbf{PDMS}$\uparrow$ \\
\midrule
DAWN & \textbf{98.2}& \textbf{95.8}& \textbf{84.2}& \textbf{100}& \textbf{95.7}& \textbf{87.9}\\
w/o World $\rightarrow$ Action & 96.6& 91.9& 78.6& 99.9& 91.6& 81.6\\
w/o Action $\rightarrow$ World & 97.3& 94.3& 80.2& 100& 92.7& 84.9\\
\bottomrule
\end{tabular}
\end{table}

\subsubsection{Does DAWN Need Full World Rollout?}


\begin{wraptable}[12]{r}{0.46\textwidth}
\centering
\vspace{-0.35cm}
\caption{Further analysis on world rollout horizon. $T_w$ denotes the latent world rollout horizon, $H_a$ denotes the action horizon, and w/o Int. reports the result without interactive refinement.}
\label{tab:rollout_horizon}
\small
\setlength{\tabcolsep}{4.3pt}
\begin{tabular}{c|c|c|c|r}
\toprule
 $\boldsymbol{T_w}$ & $\boldsymbol{H_a}$ & \textbf{PDMS}$\uparrow$& \textbf{w/o Int.}$\uparrow$ & \textbf{Lat. (ms)}$\downarrow$\\
\midrule
 0s & 4s & 82.8 & 82.8 & 331.253\\
 1s & 4s & 84.7 & 83.9 & 503.261\\
 2s & 4s & 87.3 & 84.3 & 690.540\\
 3s & 4s & 87.5 & 84.6 & 849.512\\
 4s & 4s & 87.9 & 85.2 & 1067.975\\
\bottomrule
\end{tabular}
\end{wraptable}

Although our main experiments use a 4s rollout to report the strongest configuration, these ablations examine how the rollout horizon affects DAWN. As shown in Table~\ref{tab:rollout_horizon}, zero rollout performs clearly worse, indicating that explicit future evolution remains useful in complex driving scenes. However, most of the gain appears with a shorter 2--3s latent rollout, which already approaches the full 4s result. Meanwhile, latency increases steadily as the rollout horizon becomes longer, showing the expected accuracy--efficiency trade-off. This suggests that the world branch does not need to behave as a full future simulator. Instead, it provides a compact, action-relevant dynamic hypothesis for long-horizon trajectory generation. The w/o Int. column further shows that rollout alone is not enough: predicted future latents are more useful when they interact with action refinement. This supports the WAIM view that the value of world modeling lies in interactive world-action inference, rather than passive future prediction alone.

\begin{figure*}[t]
    \centering
    \includegraphics[width=0.99\linewidth]{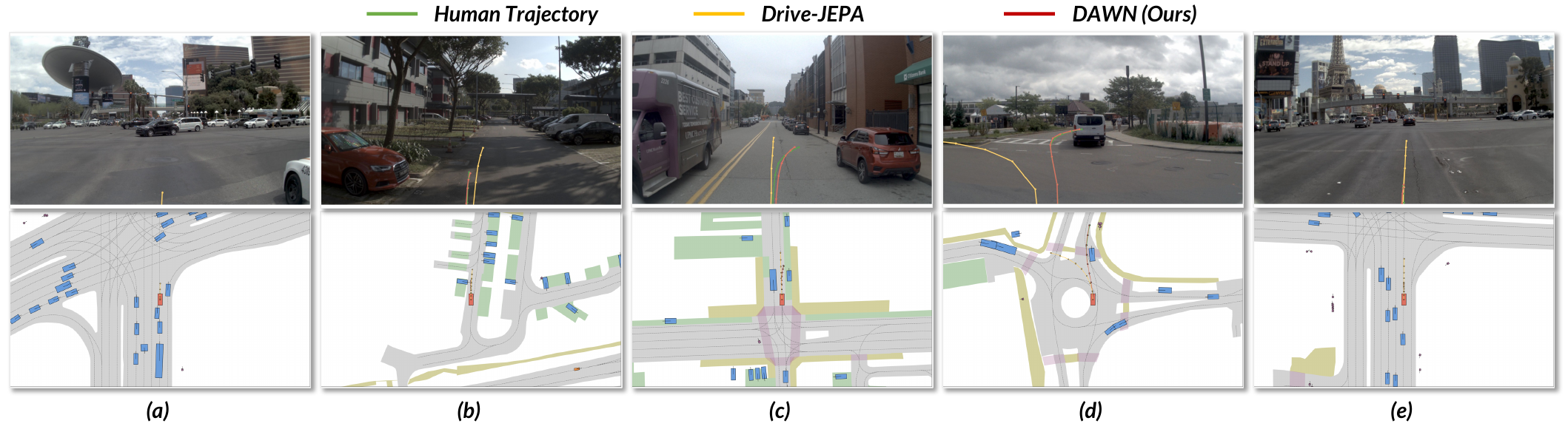}
    \caption{\textbf{Qualitative planning results.} We compare human trajectories, Drive-JEPA, and DAWN in five representative driving scenarios. The top row shows front-view observations, and the bottom row shows the corresponding BEV visualization. DAWN produces trajectories that better follow road geometry and remain visually consistent with human driving behavior in complex intersections, narrow streets, and curved junctions.}
\label{fig:qualitative}
\end{figure*}

\subsubsection{Can DAWN Generate Plausible and Safe Trajectories?}

Fig.~\ref{fig:qualitative} visualizes representative planning results in diverse urban scenarios, including wide intersections, narrow streets, constrained roads with nearby vehicles, curved junctions, and dense city intersections. In these cases, DAWN generates trajectories that are visually consistent with the human trajectory and the local road topology. For example, in the narrow-street and constrained-vehicle scenes, DAWN keeps the trajectory within the feasible driving corridor instead of drifting toward parked vehicles or road boundaries. In the curved-junction case, DAWN follows the road geometry more naturally, showing that the model can adapt its action hypothesis to non-straight layouts. These qualitative results complement the quantitative evaluation and suggest that DAWN can produce plausible and safety-aware trajectories in interactive driving scenes.
\section{Related Work}

\subsection{World Action Models}
World models provide a general framework for embodied intelligence~\cite{li2025comprehensive}. By learning how the environment evolves over time, they support prediction, planning, and decision making~\cite{hu2023gaia,li2026wildworld,seong2025vla,jia2023adriver}. Recent advances show that large-scale self-supervised learning can further strengthen world understanding. V-JEPA 2~\cite{assran2025v} learns predictive representations from internet-scale videos and supports downstream planning via latent action-conditioned models. However, they mainly perform passive prediction and treat actions as external inputs, whereas World Action Models (WAMs) jointly model future world states and ego-actions in a unified latent space~\cite{liu2026unidwm,zhang2025future}.WAM-Flow~\cite{xu2025wam} advances this by casting trajectory planning as discrete flow matching for efficient parallel refinement, while Latent-WAM~\cite{wang2026latent} introduces spatially-aware compressive encoders to extract planning-centric tokens. Beyond structural design, DreamZero~\cite{ye2026world} leverages video diffusion backbones to learn complex physical dynamics, whereas Fast-WAM~\cite{yuan2026fast}  suggests that performance gains primarily stem from video co-training rather than inference-time imagination. To enhance grounding, Percept-WAM~\cite{han2025percept} unifies 2D/3D perception tokens directly into the action space. However, existing WAMs largely rely on one-pass prediction or feedforward generation, lacking iterative reasoning mechanisms to refine both world and action jointly.

\subsection{End-to-end Autonomous Driving}
End-to-end autonomous driving maps raw sensor inputs directly to actions to simplify traditional modular pipelines~\cite{tang2026causalvad,wang2025comdrive,yang2025uncad,zhou2026opendrivevla,han2025dme,chen2025int2planner,li2025generative}. To improve robustness, UniAD~\cite{hu2023planning} unifies full-stack tasks into a single planning-optimized network, while VADv2~\cite{chen2024vadv2} introduces probabilistic planning over discretized tokens to handle environmental uncertainty. SparseDrive~\cite{sun2025sparsedrive} proposes query-centric alternatives to dense grids for higher efficiency. ReAL-AD~\cite{lu2025real} further introduces a reasoning-augmented learning framework that decomposes driving into strategy, decision, and operation levels, improving both interpretability and human-like hierarchical reasoning. Recently, Drive-JEPA~\cite{wang2026drive} adapts the Video Joint-Embedding Predictive Architecture with multimodal trajectory distillation to learn planning-aligned representations from large-scale videos. Other works like Orion~\cite{fu2025orion} and UniDriveVLA~\cite{li2026unidrivevla} incorporate MLLMs to bridge semantic reasoning with precision action generation through instruction tuning.

\subsection{Driving World Models}
Driving World Models (DWMs) focus on modeling how the environment evolves over time, typically through forward prediction of scene dynamics~\cite{zheng2025world4drive,gao2024vista,zhao2025drivedreamer4d,yang2026worldrft,li2025driverse}. DWMs serve as internal simulators that learn to internalize the principles governing scene evolution. Video models like GAIA-1~\cite{hu2023gaia}, Drive-WM~\cite{wang2024driving}, and Drive-JEPA~\cite{wang2026drive} construct predictive world representations from visual histories, with Drive-JEPA further combining video pretraining and trajectory distillation to support end-to-end planning. UniFuture~\cite{liang2025seeing} and HERMES~\cite{zhou2025hermes} enforce 4D geometric constraints. Recent methods move beyond visual forecasting toward policy-aware simulation: Uni-World VLA~\cite{liu2026uni} interleaves future frame prediction and trajectory planning to form a closed-loop interaction. For enhanced complexity, SGDrive~\cite{li2026sgdrive} and Infinite-World~\cite{wu2026infinite} introduce hierarchical cognition and memory to scale simulations to long horizons. However, most prior DWMs still treat world prediction as a passive backdrop for planning.

\section{Conclusion}
\label{sec:conclusion}

We introduced \emph{World-Action Interactive Models} (WAIMs), a perspective in which future world states and actions are inferred as coupled variables rather than produced by decoupled pipelines. Based on this idea, we proposed DAWN, a latent generative model that couples a World Predictor with a World-Conditioned Action Denoiser through short explicit latent rollout. Experiments show that this design improves planning quality, interactive safety, and trajectory smoothness while remaining efficient at inference time. We hope this work encourages further exploration of interactive world-action generation for more actionable autonomous systems.

\section{Acknowledgments}

This work was supported in part by the Research and Application of Key Technologies for L4 End-to-End Autonomous Driving Based on Multi-modal Large Language Models under Grant 202423dl2050005, and in part by Research and Application of the Next-Generation General-Purpose Intelligent Robot Brain (Robo-GPT) under Grant 2024zd01.

\clearpage

\bibliographystyle{plainnat}
\bibliography{main}


\clearpage
\beginappendix
\section{Limitations}
\label{app:limitations}


This work has limitations at both the WAIM formulation level and the DAWN instantiation level. At the formulation level, WAIM assumes that future world states and future actions should be inferred as coupled variables. This is suitable for action-contingent and interactive decision-making problems, but it may be unnecessary in simpler settings where a direct policy or a zero-rollout WAM already provides a better efficiency--performance trade-off. In addition, our current WAIM formulation does not provide formal convergence or safety guarantees for the recursive interaction between world and action hypotheses.

At the instantiation level, DAWN realizes WAIM through a short latent world rollout. This design improves efficiency, but may be insufficient for scenarios requiring long-range anticipation or extended multi-agent interaction. Since DAWN performs world-action interaction in a compact latent space, the learned future representation is also less interpretable than explicit scene-level predictions, making it difficult to diagnose whether rare safety-critical cues are preserved. Finally, both WAIM and DAWN remain data-driven and depend on the coverage of pretraining and downstream driving datasets. The reported benchmark gains should therefore be interpreted as improved performance under standard evaluation protocols, not as evidence of deployment readiness.
\section{Broader Impact}
\label{app:broader_impact}

This work may have positive impact by improving how autonomous systems reason about future consequences before acting. In autonomous driving and other embodied settings, stronger world-action models could support safer planning, smoother interaction, and more efficient future reasoning. At the same time, these capabilities also introduce risks. More capable decision models may encourage over-trust in partially validated autonomy, and uneven generalization across regions, environments, or traffic patterns could create unfair or unsafe outcomes. In addition, large-scale driving video data may raise privacy concerns, and models of this kind could be misused in aggressive autonomous navigation or surveillance settings. Although DAWN reduces inference-time cost relative to full pixel-space rollout, its multi-stage training pipeline is still computationally intensive. We therefore view this work as a research step toward safer and more actionable world models, not as a justification for unrestricted real-world deployment.

\section{More Experiments Details}
\label{app:details}

\subsection{Datasets and Metrics}
\label{app:datasets_metrics}

We evaluate DAWN on four autonomous driving benchmarks covering both open-loop and closed-loop settings: NAVSIM v1, NAVSIM v2, and nuScenes.

\noindent \textbf{NAVSIM v1.} NAVSIM is a real-world benchmark based on large-scale driving data and evaluates planning quality through simulator-based rule metrics. Following the standard protocol, we report NC (No-at-fault Collisions), DAC (Drivable Area Compliance), EP (Ego Progress), C (Comfort), and TTC (Time-to-Collision), together with the aggregated PDMS score.

\noindent \textbf{NAVSIM v2.} Compared with NAVSIM v1, NAVSIM v2 strengthens the evaluation by extending PDMS to EPDMS and introducing richer rule-compliance and comfort measures. In addition to NC, DAC, EP, and TTC, it reports DDC (Driving Direction Compliance), TL (Traffic Light Compliance), LK (Lane Keeping), HC (History Comfort), and EC (Extended Comfort). We use EPDMS as the main aggregate metric.


\noindent \textbf{nuScenes.} On the nuScenes planning benchmark, we follow the standard end-to-end planning protocol and report trajectory L2 error and Collision Rate at 1\,s, 2\,s, and 3\,s, as well as the average over the horizon. These metrics measure motion accuracy and safety, respectively.

For NAVSIM, higher values indicate better performance for all reported metrics. For nuScenes, lower values are better for both L2 error and collision rate.

\subsection{Detail Experimental Settings}
\label{app:set}

We provide additional implementation details for reproducibility. Input clips are sampled at 2\,Hz with a crop size of $512\times256$. We use a ViT-Large V-JEPA 2 backbone with patch size 16 and tubelet size 2. The model observes 4 frames and predicts future latent states over 12 target frames. Training uses bfloat16 mixed precision and scaled dot-product attention when available.

The Auto-Encoder Resampler compresses dense encoder tokens into 16 latent tokens. It uses 16 attention heads, a 4-layer encoder, a 2-layer decoder, MLP ratio 4.0, learnable positional embeddings, and no dropout. During resampler training, we attach an auxiliary diffusion planner head with the same DiT-style configuration as the World-Conditioned Action Denoiser, which encourages the compressed tokens to preserve action-relevant information. The World Predictor is a causal Transformer with 12 layers, embedding dimension 384, 12 attention heads, RoPE positional encoding, and activation checkpointing.

The World-Conditioned Action Denoiser is implemented as a diffusion planner with a DiT-style backbone. It uses hidden dimension 384, 12 layers, 12 attention heads, MLP ratio 4.0, and no dropout. Compared with our initial diffusion planner block, we modify the DiT block to more closely follow the original adaLN-Zero design: the timestep/status conditioning vector modulates not only the self-attention and MLP branches, but also the cross-attention branch to the latent world tokens. Specifically, each block predicts shift, scale, and gate parameters for self-attention, cross-attention, and MLP residual branches. We also remove the additional unmodulated MLP after cross-attention, so that the block follows a cleaner sequence of modulated self-attention, modulated cross-attention, and modulated feed-forward update. This design makes the action denoising process more consistently conditioned on diffusion time, ego status, and latent world context. The denoiser predicts multimodal trajectory hypotheses with 6 modes/samples and uses 5 DPM-Solver++ sampling steps at inference. We represent trajectories with per-pose tokens at a temporal interval of 0.5\,s. The diffusion objective combines classification, regression, velocity, and yaw losses, with the velocity and yaw terms weighted by 0.5.

For optimization, we train for 150 epochs with a peak learning rate of $1\times10^{-4}$, initial learning rate $5\times10^{-5}$, weight decay 0.04, and 8 warmup epochs. EMA momentum is increased from 0.996 to 0.999 during training. Our large-scale experiments are trained on 80 NVIDIA A100 GPUs. We also verified that the training pipeline can be launched on a single RTX 4090 for debugging and small-scale runs, although the reported full-scale results use A100 training.

\section{More Quantitative Results}
\label{app:exp}

\subsection{Comparison with existing SOTA Methods on other Datasets}

\begin{table}[H]
\centering
\caption{Quantitative comparisons on NAVSIM v2. We report perception-based baselines for reference and include DAWN under the same official NAVSIM v2 evaluation protocol. Higher values are better for all metrics, and EPDMS is the aggregate score.}
\label{tab:navsim_v2}
\resizebox{\linewidth}{!}{%
\begin{tabular}{l|cccc|ccccc|c}
\toprule
\textbf{Method} & \textbf{NC}$\uparrow$ & \textbf{DAC}$\uparrow$ & \textbf{DDC}$\uparrow$ & \textbf{TL}$\uparrow$ & \textbf{EP}$\uparrow$ & \textbf{TTC}$\uparrow$ & \textbf{LK}$\uparrow$ & \textbf{HC}$\uparrow$ & \textbf{EC}$\uparrow$ & \textbf{EPDMS}$\uparrow$ \\
\hline
Transfuser~\cite{chitta2022transfuser} & 96.9 & 89.9 & 97.8 & 99.7 & 87.1 & 95.4 & 92.7 & 98.3 & 87.2 & 76.7 \\
Hydra-MDP++~\cite{li2025hydra} & 98.5 & 98.5 & 99.5 & 99.7 & 87.4 & 97.9 & 95.8 & 98.2 & 75.7 & 85.6 \\
iPad~\cite{guo2025ipad} & 98.7 & 98.0 & 98.9 & 99.8 & 86.6 & 98.3 & 97.2 & 98.3 & 74.6 & 85.8 \\
DriveSuprim~\cite{yao2026drivesuprim} & 98.4 & 98.6 & 99.6 & 99.8 & 90.5 & 97.8 & 97.0 & 98.3 & 78.6 & 87.1 \\
Drive-JEPA~\cite{wang2026drive} & 98.4 & 98.6 & 99.1 & 99.8 & 88.4 & 97.8 & 97.6 & 97.9 & 84.8 & 87.8 \\
\hline \rowcolor{lightblue} 
\textbf{DAWN (Ours)}&97.3&92.0&99.1&99.7&87.4&96.6&96.0&98.3&	85.5&83.2\\
\bottomrule
\end{tabular}%
}
\end{table}

\noindent\textbf{Results on NAVSIM v2.}
Table~\ref{tab:navsim_v2} reports NAVSIM v2 results, where we compare DAWN with representative perception-based baselines under the official evaluation protocol. DAWN achieves the best extended-comfort score and competitive lane-keeping performance, while maintaining strong traffic-light compliance and history-comfort scores, indicating smooth and stable trajectories. Its aggregate EPDMS is lower than the strongest baselines, mainly due to weaker drivable-area compliance and collision-related scores, suggesting that strict rule compliance remains a direction for further improvement.

\subsection{Detailed Ablation Studies}
\begin{table}[H]
\centering
\caption{Ablation study of different model components.}
\label{tab:ablation_components}
\resizebox{\textwidth}{!}{
\begin{tabular}{ccc|cccccc}
\toprule
\textbf{AE Resampler} &
\textbf{Predictor} &
\textbf{Interactive} &
\textbf{NC$\uparrow$} &
\textbf{DAC$\uparrow$} &
\textbf{EP$\uparrow$} &
\textbf{TTC$\uparrow$} &
\textbf{C$\uparrow$} &
\textbf{PDMS$\uparrow$} \\
\midrule
           &             &             & 97.1& 92.2& 78.8& 100& 91.5& 82.9\\
\checkmark &             &             & 97.2& 92.2& 78.7& 100& 91.7& 82.8\\
\checkmark & \checkmark  &             & 97.4& 94.3& 80.4& 100& 91.5& 85.2\\
\checkmark & \checkmark  & \checkmark  & \textbf{98.2}& \textbf{95.8}& \textbf{84.2}& \textbf{100}& \textbf{95.8}& \textbf{87.9}\\
\bottomrule
\end{tabular}
}
\end{table}

\noindent\textbf{Detailed ablation on key components.}
The components ablation in Table~\ref{tab:ablation_components} reveals a clear progression in performance as we incrementally introduce the key modules of DAWN. Using only a basic backbone without the resampler yields limited performance, and adding the AE Resampler alone brings negligible improvement, indicating that compact latent compression by itself is insufficient for planning. Introducing the World Predictor leads to a noticeable gain across all metrics, improving PDMS from 82.8 to 85.2, which suggests that explicit latent future rollout provides useful structural guidance for downstream action generation. The largest improvement comes from enabling the interactive design, where world prediction and action generation are coupled. This further boosts PDMS to 87.9 and consistently improves all safety-related metrics (e.g., TTC and DAC), demonstrating that the benefit is not only from better world modeling, but from allowing world and action hypotheses to be jointly refined during inference. Overall, the results validate that each component contributes differently: the resampler provides a compact representation, the predictor introduces temporal reasoning, and the interaction mechanism is the key factor that translates these into improved planning performance.

\begin{table}[H]
\centering
\caption{Ablation study on the number of interactive rounds.}
\label{tab:refine_rounds_ablation}
\begin{tabular}{c|cccccc}
\toprule
\textbf{\# Rounds} & \textbf{NC}$\uparrow$ & \textbf{DAC}$\uparrow$ & \textbf{EP}$\uparrow$ & \textbf{C}$\uparrow$ & \textbf{TTC}$\uparrow$ & \textbf{PDMS}$\uparrow$  \\
\hline
1 & 97.4& 94.3& 80.4& 100& 91.5& 85.2 \\
2 & 97.8& 95.1& 81.6& 100& 94.1& 86.4 \\
3 & 98.1& 95.6& 82.8& 100& 95.6& 86.9 \\
4 & \textbf{98.2}& \textbf{95.8}& \textbf{84.2}& \textbf{100}& \textbf{95.8}& \textbf{87.9} \\
5 & 98.1& 95.4& 83.9& 100& 95.7& 87.2 \\
6 & 98& 95.6& 82.8& 100& 95.6& 86.9 \\
\bottomrule
\end{tabular}
\end{table}

\noindent\textbf{Ablation on interactive rounds.}
Table~\ref{tab:refine_rounds_ablation} provides the full numerical results for different numbers of interactive rounds. Increasing the number of rounds from 1 to 4 consistently improves the aggregate PDMS score from 85.2 to 87.9, with simultaneous gains in NC, DAC, EP, and TTC. This confirms that the recursive interaction between the World Predictor and the World-Conditioned Action Denoiser is beneficial, rather than being a one-step conditioning effect. Beyond 4 rounds, the performance no longer improves: PDMS drops to 87.2 with 5 rounds and 86.9 with 6 rounds. We therefore set the default number of interactive rounds to 4 in all main experiments.

\begin{table}[H]
\centering
\caption{Ablation study on the number of latent tokens in the Auto-Encoder Resampler.}
\label{tab:ae_token_ablation}
\begin{tabular}{c|ccccccr}
\toprule
\textbf{\# Tokens} & \textbf{NC}$\uparrow$ & \textbf{DAC}$\uparrow$ & \textbf{EP}$\uparrow$ & \textbf{C}$\uparrow$ & \textbf{TTC}$\uparrow$ & \textbf{PDMS}$\uparrow$ & \textbf{Latency}$\downarrow$ \\
\hline
16  & 97.2& 92.2& 78.7& 100& 91.7& 82.8& 331.253\\
64  & 97.2& 92.4& 78.8& 100& 91.7& 83.2& 963.645\\
\bottomrule
\end{tabular}
\end{table}


\noindent\textbf{Ablation on resampler latent tokens.}
The ablation on the number of latent tokens in Table~\ref{tab:ae_token_ablation} highlights a clear trade-off between latent capacity and computational efficiency. While expanding the output from 16 to 64 tokens slightly improves the aggregate PDMS from 82.8 to 83.2, alongside minor gains in Drivable Area Compliance (DAC) and Ego Progress (EP), it also increases inference latency by more than 3$\times$ (from 331.3 ms to 963.6 ms). This suggests that a compact 16-token representation is already highly effective at capturing the essential scene structure for planning, and that the marginal performance gains from expanding the token capacity do not justify the substantial computational overhead. This finding aligns with the core design goal of the AE Resampler, which is to distill the scene into a minimal yet informative latent space. In the context of DAWN, effective planning relies more on structured world-action interaction than on raw high-dimensional latent capacity. Therefore, maintaining a compact representation provides the optimal balance, preserving strong planning accuracy while retaining the efficiency required for practical deployment.

\begin{figure*}[t]
    \centering
    \includegraphics[width=0.60\linewidth]{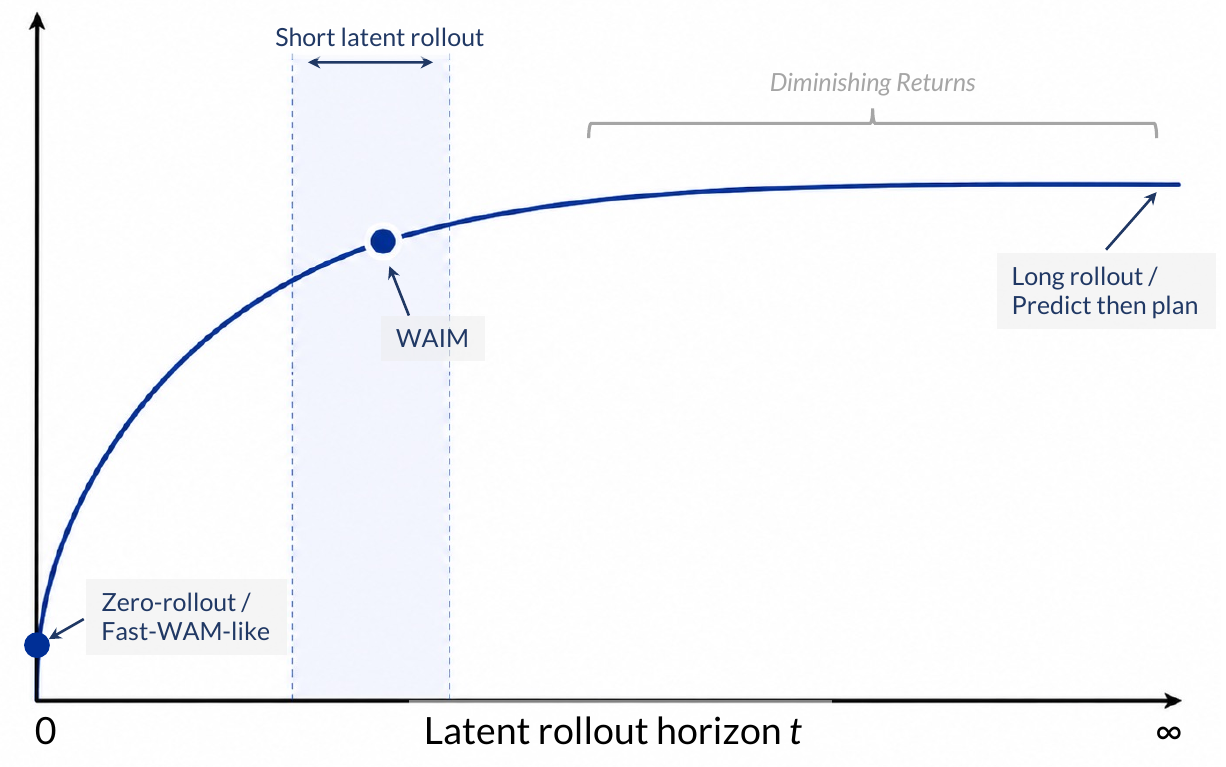}
    \caption{Illustration of the latent world rollout design space. Zero-rollout methods such as Fast-WAM occupy the left endpoint, full predict-then-plan methods occupy the right endpoint, and DAWN targets a short-rollout regime in between, where compact future evolution provides useful foresight without full-horizon rollout.}
    \label{fig:rollout_continuum}
\end{figure*}

\noindent\textbf{Latent rollout as a continuum.}
Fig.~\ref{fig:rollout_continuum} summarizes the design space explored in our rollout ablation. Zero-rollout methods, such as Fast-WAM-like variants, rely entirely on latent representations learned during training and do not explicitly evolve the world at inference time. At the other extreme, full predict-then-plan methods roll out the future over the entire action horizon, but this is not always necessary for planning. The observed trend suggests diminishing returns as the rollout horizon grows: most of the performance gain appears once the model is allowed to reason over a short latent future, while extending rollout further yields smaller additional benefit. This supports the WAIM perspective that the useful future for decision making is often a compact, action-relevant hypothesis rather than a full reconstruction of the future scene. DAWN is therefore positioned in the short-rollout regime, where explicit world evolution is retained but kept efficient enough to remain practical.

\clearpage

\section{More Qualitative Results}
\label{app:vis}

\subsection{Planning Results}
\begin{figure}[H]
    \centering
    \includegraphics[width=0.79\linewidth]{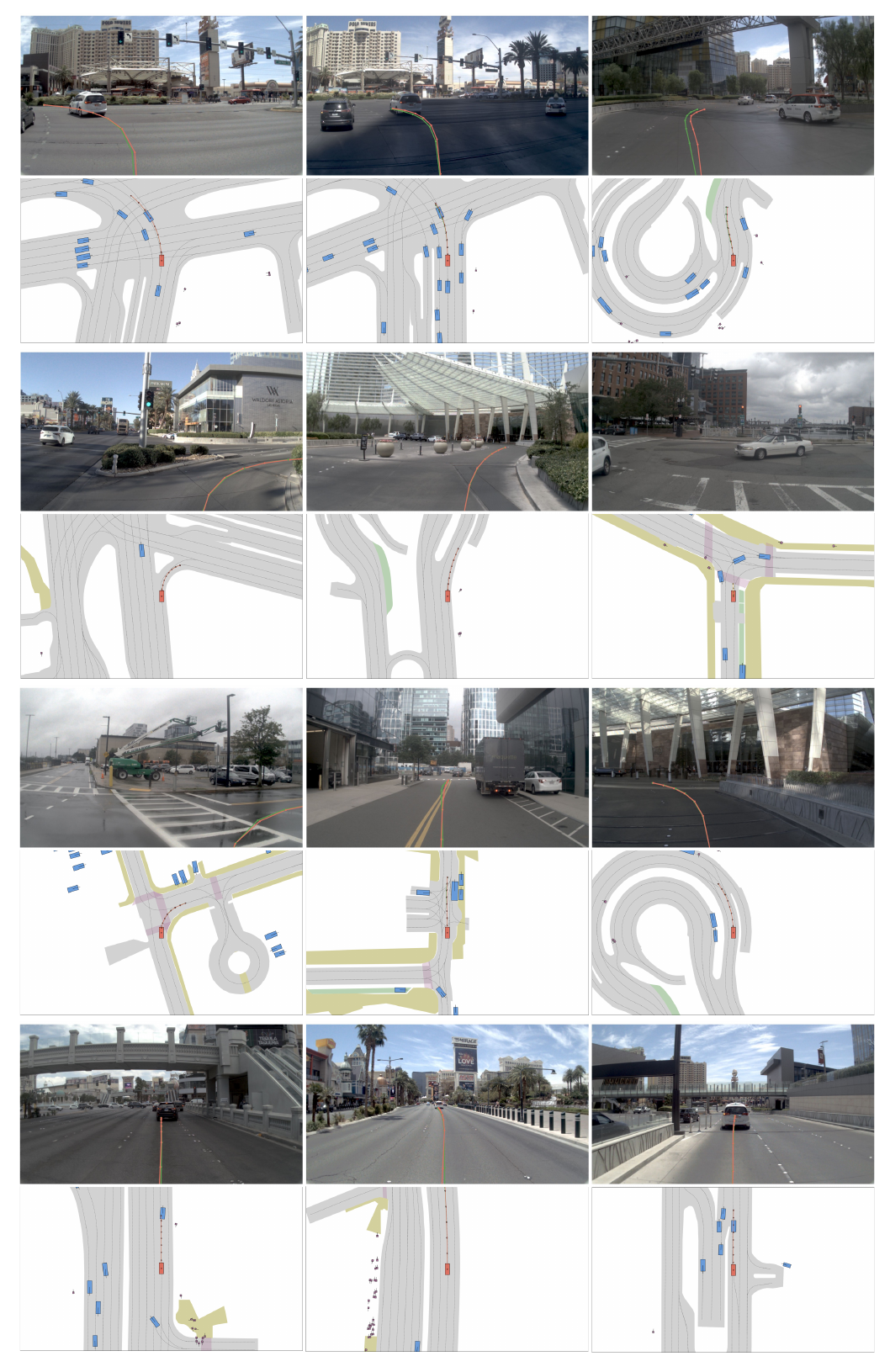}
    \caption{More qualitative results of planning.}
    \label{app:vis}
\end{figure}
\clearpage

\subsection{Prediction Results}
\begin{figure}[H]
    \centering
    \includegraphics[width=0.85\linewidth]{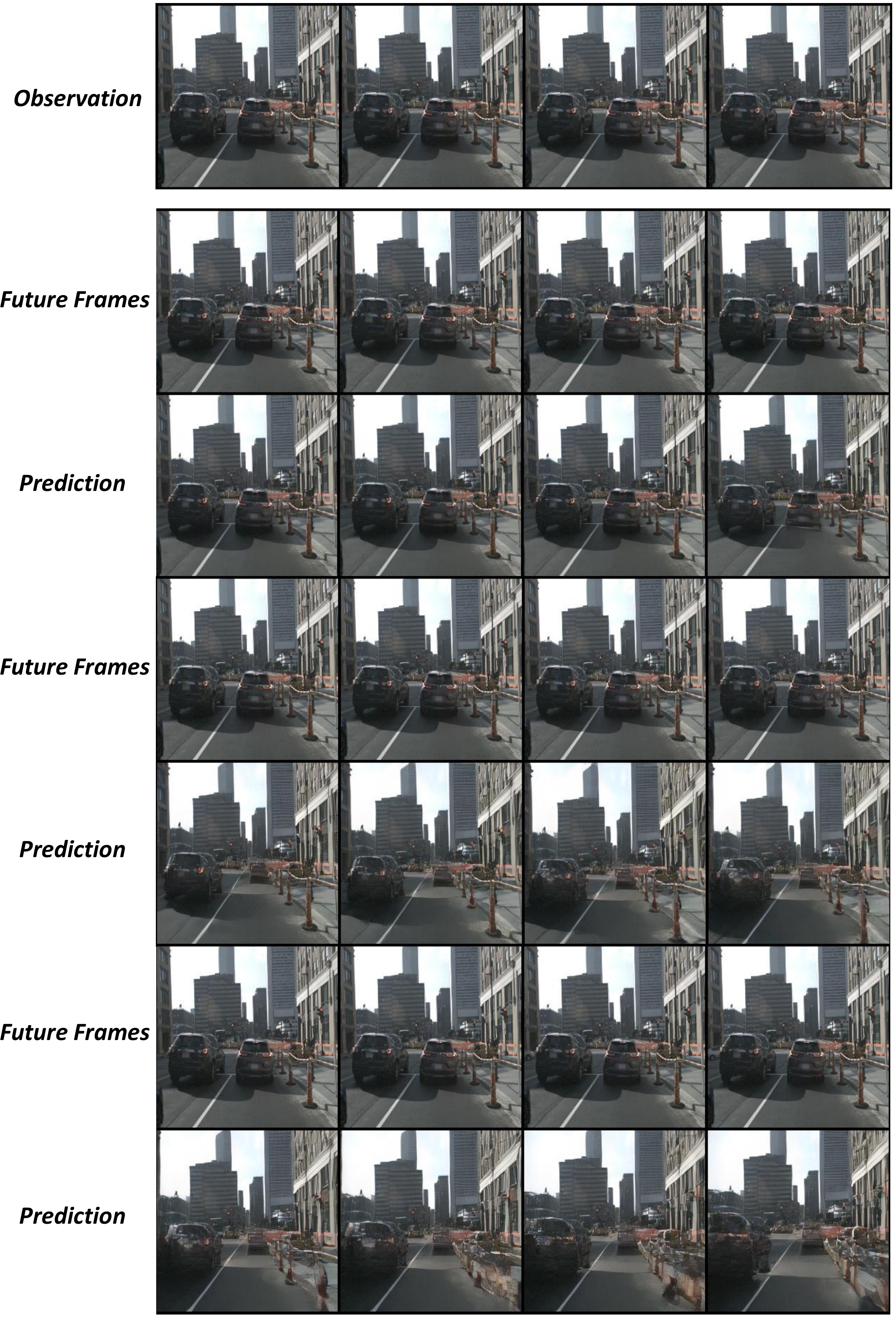}
    \caption{More qualitative results of prediction.}
    \label{app:prediction1}
\end{figure}

\clearpage

\begin{figure}[H]
    \centering
    \includegraphics[width=0.85\linewidth]{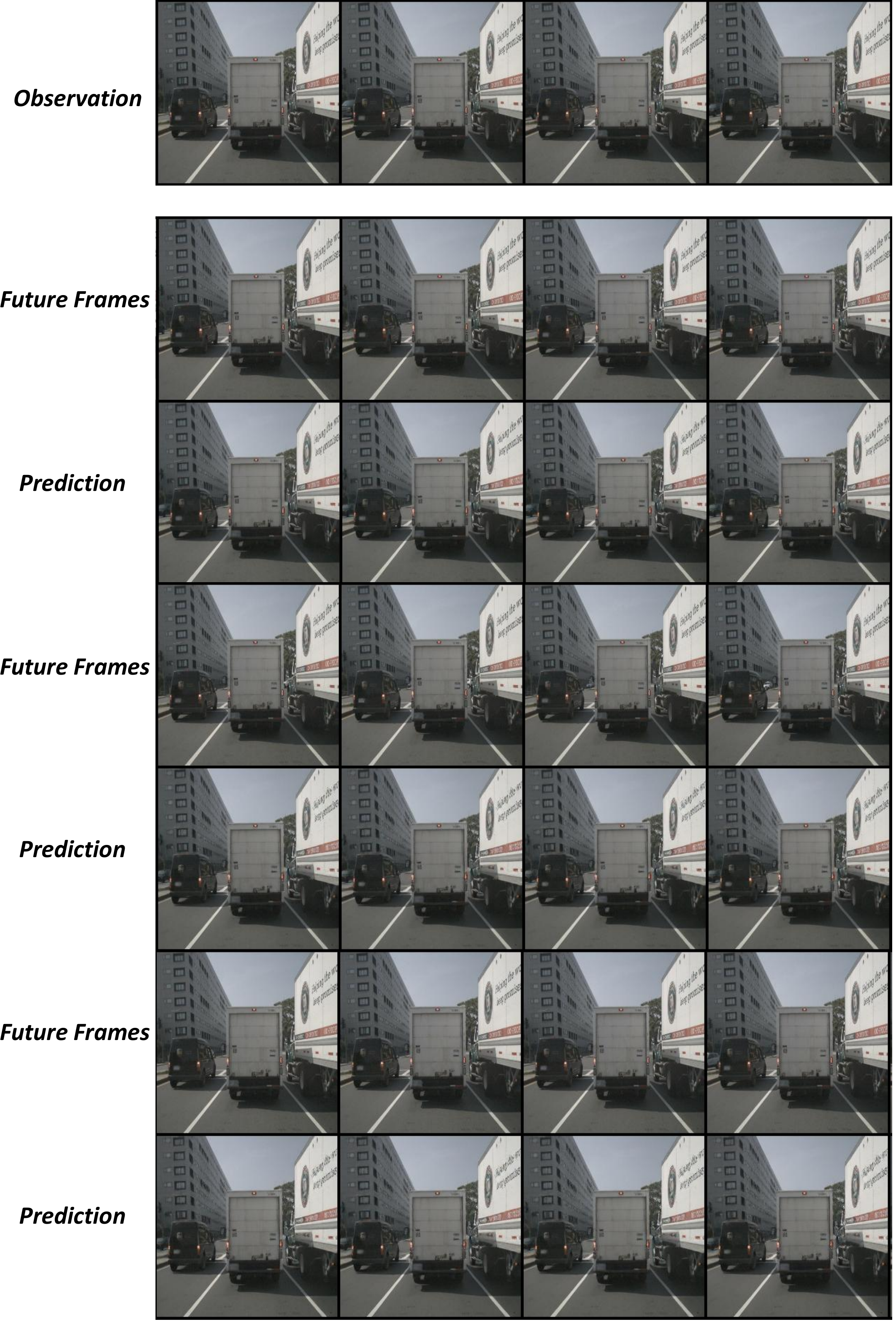}
    \caption{More qualitative results of prediction.}
    \label{app:prediction2}
\end{figure}
\clearpage

\begin{figure}[H]
    \centering
    \includegraphics[width=0.85\linewidth]{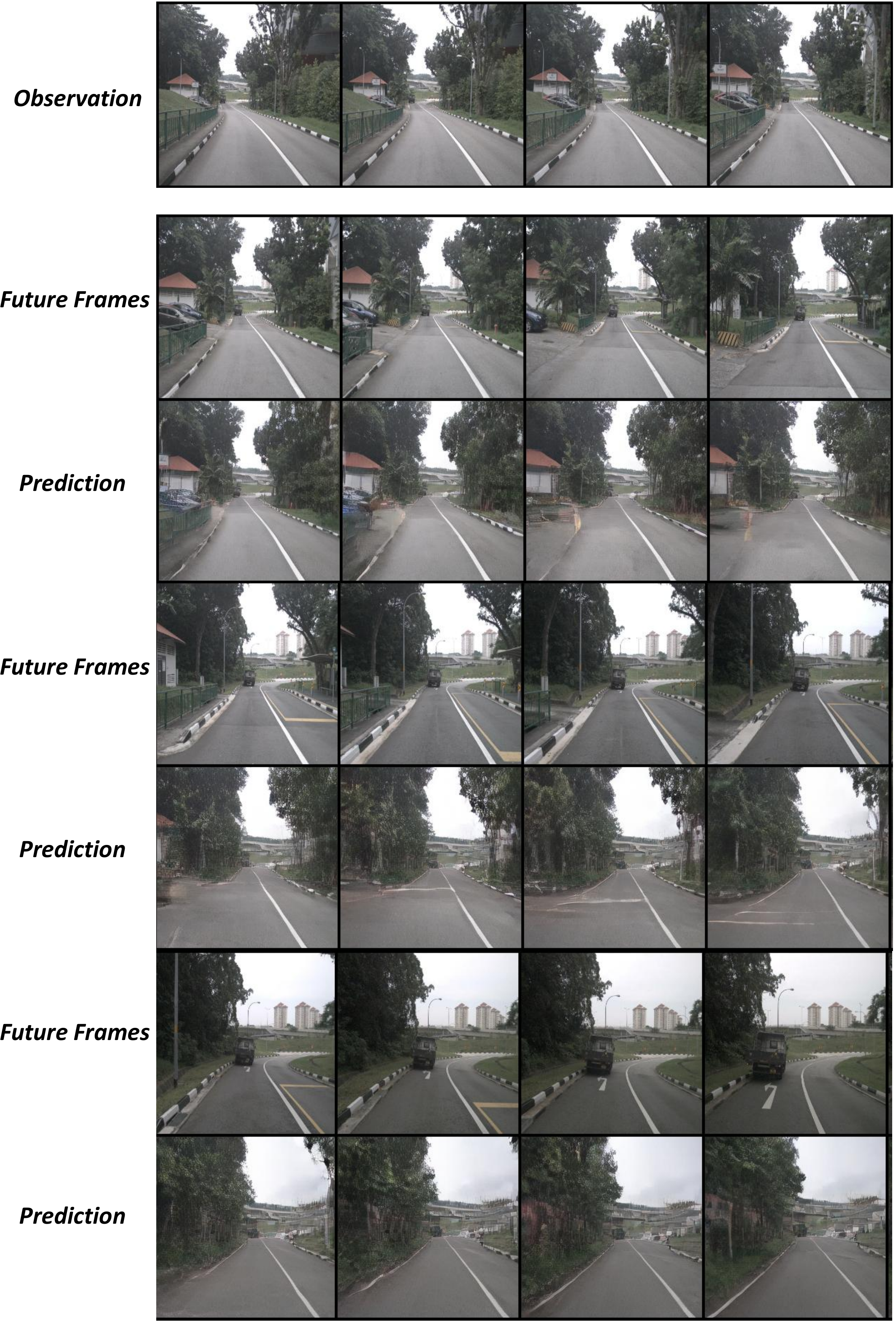}
    \caption{More qualitative results of prediction.}
    \label{app:prediction3}
\end{figure}

\subsection{Featuremap}

\begin{figure}[H]
    \centering
    \includegraphics[width=0.9\linewidth]{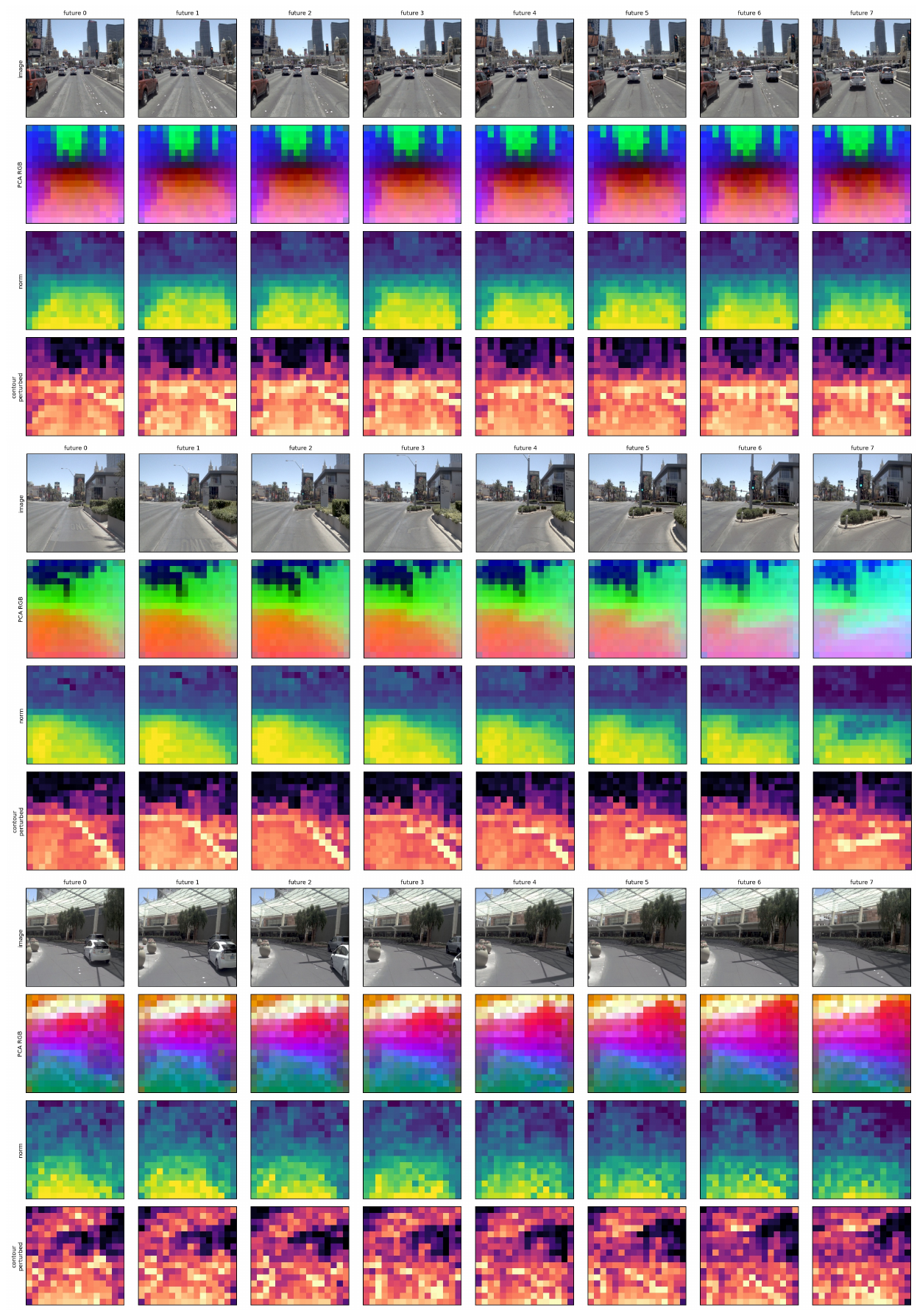}
    \caption{More qualitative results of feature.}
    \label{app:feature1}
\end{figure}

\clearpage

\begin{figure}[H]
    \centering
    \includegraphics[width=0.9\linewidth]{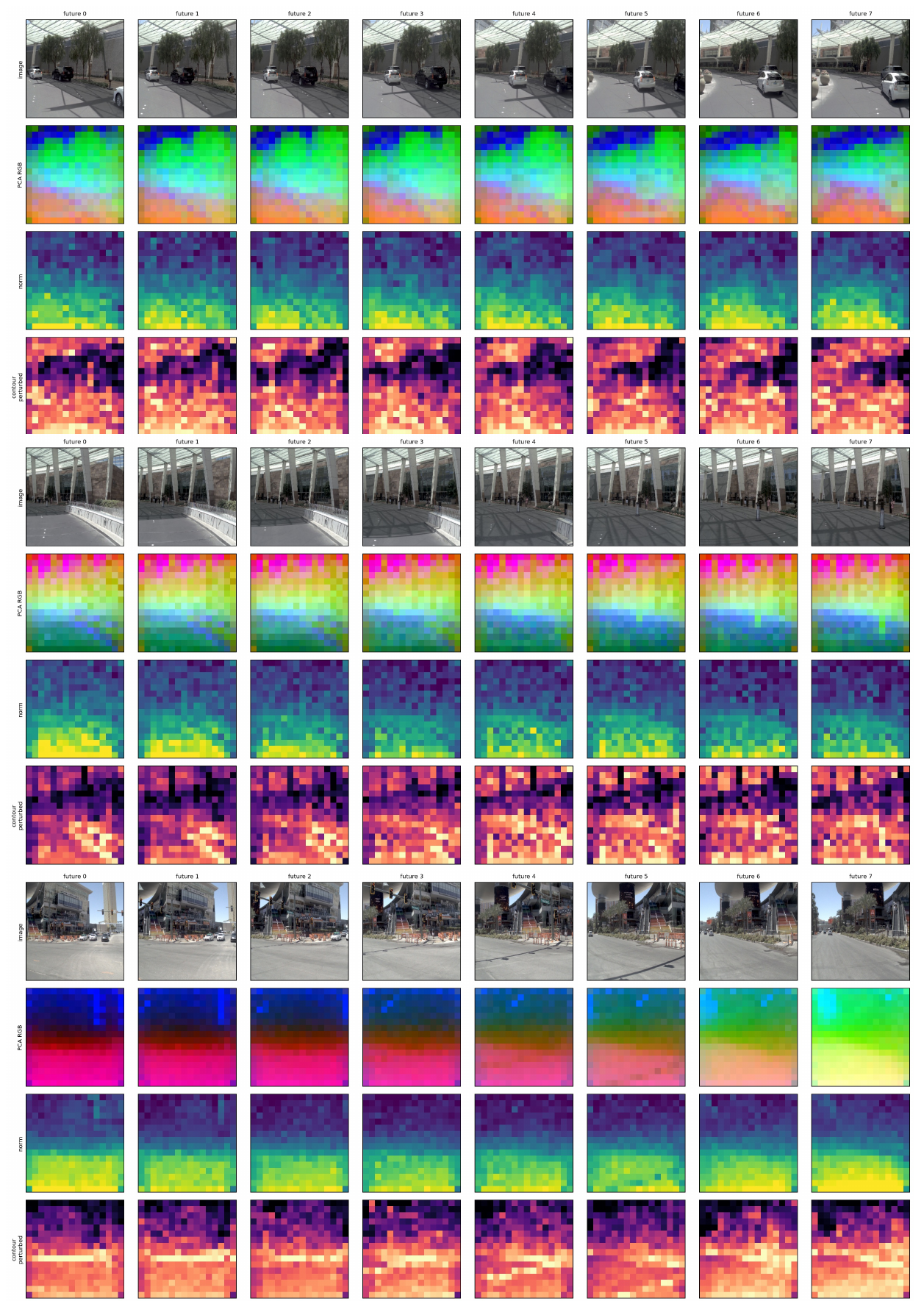}
    \caption{More qualitative results of feature.}
    \label{app:feature2}
\end{figure}

\clearpage

\section{Pseudo Code of DAWN}
\label{app:alg}


\begin{algorithm}[H]
\caption{DAWN Training}
\label{alg:dawn_training}
\begin{algorithmic}[1]
\Require Pretraining data $\mathcal{D}_{\mathrm{pre}}$, task data $\mathcal{D}_{\mathrm{task}}$
\Require Encoders $E_{\mathrm{stu}},E_{\mathrm{tea}}$, resamplers $R_{\mathrm{stu}},R_{\mathrm{tea}}$
\Require World Predictor $P_{\theta}$, Action Denoiser $G_{\phi}$, Action Head $H_{\mathrm{act}}$
\Ensure Trained DAWN

\State \textbf{Stage 1: Vision pretraining.}
\State Pretrain $E_{\mathrm{stu}}$ on unified driving videos from $\mathcal{D}_{\mathrm{pre}}$; update $E_{\mathrm{tea}}$ by EMA.

\State \textbf{Stage 2: Resampler training.}
\State Train $R_{\mathrm{stu}}$ as a token-space autoencoder on dense encoder tokens $E_{\mathrm{stu}}(o)$.

\State \textbf{Stage 3: World predictor training.}
\For{each $(o,l,o^{+}) \in \mathcal{D}_{\mathrm{task}}$}
\State $z \leftarrow R_{\mathrm{stu}}(E_{\mathrm{stu}}(o))$, \quad
$z_{\mathrm{tar}} \leftarrow R_{\mathrm{tea}}(E_{\mathrm{tea}}(o^{+}))$
\State $\hat{z}_{\mathrm{fut}} \leftarrow P_{\theta}(z,c)$
\State Update $P_{\theta}$ by minimizing
$\mathcal{L}_{\mathrm{WM}} = d(\hat{z}_{\mathrm{fut}}, z_{\mathrm{tar}})$
\EndFor

\State \textbf{Stage 4: Joint world-action training.}
\State Initialize $P_{\theta}$ from Stage 3 and attach $G_{\phi}$ and $H_{\mathrm{act}}$
\For{each $(o,l,o^{+},\tau^{\star}) \in \mathcal{D}_{\mathrm{task}}$}
\State $z \leftarrow R_{\mathrm{stu}}(E_{\mathrm{stu}}(o))$, \quad
$z_{\mathrm{tar}} \leftarrow R_{\mathrm{tea}}(E_{\mathrm{tea}}(o^{+}))$
\State $a_{1:H}^{(0)} \leftarrow G_{\phi}(q_{\mathrm{prop}},c,z)$
\For{$r=0$ to $R-1$}
\State $z_{\mathrm{fut}}^{(r)} \leftarrow P_{\theta}(z,c,a_{1:H}^{(r)})$
\State $a_{1:H}^{(r+1)} \leftarrow G_{\phi}(q_{\mathrm{ref}}^{(r)},c,z_{\mathrm{fut}}^{(r)},a_{1:H}^{(r)})$
\EndFor
\State Update $P_{\theta},G_{\phi},H_{\mathrm{act}}$ with world loss and planning loss
\EndFor

\State \textbf{return} Trained DAWN
\end{algorithmic}
\end{algorithm}


\begin{algorithm}[H]
\caption{DAWN Inference}
\label{alg:dawn_inference}
\begin{algorithmic}[1]
\Require Current observation $o$, instruction $l$, interactive rounds $K$
\Require Student Vision-Encoder $E_{\mathrm{stu}}$, Auto-Encoder Resampler $R_{\mathrm{stu}}$
\Require World Predictor $P_{\theta}$, World-Conditioned Action Denoiser $G_{\phi}$, Action Head $H_{\mathrm{act}}$
\Ensure Predicted trajectory $\hat{\tau}$

\State Extract compact latent context:
\State $z \leftarrow R_{\mathrm{stu}}(E_{\mathrm{stu}}(o))$

\State Encode non-visual conditions:
\State $c \leftarrow C(l)$

\State Initialize the action hypothesis directly from the resampler latent:
\State $a_{1:H}^{(0)} \leftarrow G_{\phi}(q_{\mathrm{init}}, c, z)$

\For{$k = 0$ to $K-1$}
\State Roll out a short latent future conditioned on the current action:
\State $z_{\mathrm{future}}^{(k+1)} \leftarrow P_{\theta}(z, c, a_{1:H}^{(k)})$
\State Refine the action hypothesis with the predicted latent future:
\State $a_{1:H}^{(k+1)} \leftarrow G_{\phi}(q_{\mathrm{ref}}^{(k)}, c, z_{\mathrm{future}}^{(k+1)}, a_{1:H}^{(k)})$
\EndFor

\State Decode the final action state:
\State $\hat{\tau} \leftarrow H_{\mathrm{act}}(a_{1:H}^{(K)})$

\State \textbf{return} $\hat{\tau}$
\end{algorithmic}
\end{algorithm}

\end{document}